%% file: main.tex
\pdfoutput=1

\documentclass[11pt]{article}

\usepackage{acl}

\usepackage{times}
\usepackage{latexsym}

\usepackage[T1]{fontenc}

\usepackage[utf8]{inputenc}

\usepackage{microtype}

\usepackage{makecell}
\usepackage{times}
\usepackage{latexsym}
\usepackage{tabularx}
\usepackage{array}
\usepackage{multirow}
\usepackage{siunitx}
\usepackage{booktabs}
\usepackage[pdftex]{graphicx}
\usepackage{enumitem}
\usepackage{xspace}
\usepackage{caption}

\usepackage{url}            
\usepackage{amsmath,amsthm,amsfonts,amssymb,bm,stmaryrd}       
\usepackage{nicefrac}       
\usepackage{enumitem}
\usepackage[noend]{algpseudocode}
\usepackage{algorithm}
\usepackage{mathtools}
\usepackage{nccmath}
\usepackage{multirow}
\usepackage{bigdelim}
\usepackage{color, colortbl}
\usepackage{xcolor}		
\definecolor{darkblue}{rgb}{0, 0, 0.5}
\hypersetup{colorlinks=true, citecolor=darkblue, linkcolor=darkblue, urlcolor=darkblue}
\usepackage{booktabs}
\usepackage{dcolumn}
\newcolumntype{d}{D{.}{.}{-1}}
\newcolumntype{z}{D{(}{\ (}{1.1}}
\usepackage{graphicx}
\usepackage{titlesec}
\usepackage{hyperref}
\usepackage{url}
\usepackage{framed}
\usepackage{tcolorbox}
\usepackage{subcaption}
\usepackage{graphicx}
\usepackage[normalem]{ulem}
\useunder{\uline}{\ul}{}
\usepackage{fontawesome}
\usepackage{calc}

\usepackage[notes=false, done=false, later=false, ]{dtrt} 
\input{macros}

\title{\vspace*{-0.5in}
{{\small \hfill \textit{ACL'23 Findings}}\\
\vspace*{.25in}} \datasetname: A Large-Scale Corpus for \\ Proposition-Level Segmentation and Entailment Recognition}

\author{Sihao Chen*$^{1,2}$ ~ Senaka Buthpitiya$^1$ ~ Alex Fabrikant$^1$ ~ Dan Roth$^2$ ~ Tal Schuster$^1$\\
\\
    $^1$Google Research \qquad $^2$University of Pennsylvania \\
    {\tt \small \{senaka,fabrikant,talschuster\}@google.com, \tt \small \{sihaoc,danroth\}@cis.upenn.edu  }
}

\begin{document}
\maketitle

\begin{abstract}
\blfootnote{* Work done as an intern at Google} 
The widely studied task of Natural Language Inference (NLI) requires a system to recognize whether one piece of text is textually entailed by another, i.e.\ whether the \emph{entirety} of its meaning can be inferred from the other. 
In current NLI datasets and models, textual entailment relations are typically defined on the sentence- or paragraph-level. 
However, even a simple sentence often contains multiple \emph{propositions}, i.e.\ distinct units of \emph{meaning} conveyed by the sentence. As these propositions can carry different truth values in the context of a given premise,
we argue for the need to recognize the textual entailment relation of each proposition in a sentence individually. 

We propose \datasetname, a corpus of over $45$K propositions annotated by expert human raters.
Our dataset structure aligns with the tasks of (1) segmenting sentences within a document to the set of propositions, and (2) classifying the entailment relation of each proposition with respect to a different yet topically-aligned document, i.e. documents describing the same event or entity. 
We establish strong baselines for the segmentation and entailment tasks. Through case studies on summary hallucination detection and document-level NLI, 
we demonstrate that our conceptual framework is potentially useful for understanding and explaining the compositionality of NLI labels.
\end{abstract}

\input{sections/intro}
\input{sections/motivation}

\input{sections/dataset}
\input{sections/baseline}
\input{sections/experiments}

\input{sections/discussion}
\input{sections/conclusion}
\input{sections/limitations}
\input{sections/acknowledgements}

\bibliography{anthology,custom}

\newpage
\appendix

\input{appendix/model_param}
\input{appendix/rater_guidelines}
\input{appendix/contradictions}
\input{appendix/openie_vs_us}
\input{appendix/xsum_examples}

\end{document}

%% file: macros.tex
\newcommand{\greencheck}{\textcolor{green}{\checkmark}}
\newcommand{\redmark}{\textcolor{red}{\ding{55}}}

\makeatletter
\def\blfootnote{\xdef\@thefnmark{}\@footnotetext}
\makeatother

\newcommand{\datasetname}{\textsc{PropSegmEnt}\xspace}

\newcommand{\newshead}{NewSHead\xspace}
\newcommand{\markstart}{\texttt{[M]}}
\newcommand{\markend}{\texttt{[/M]}}

%% file: sections/intro.tex
\section{Introduction}
Natural Language Inference (NLI), or Recognizing Textual Entailment (RTE), is the task of determining whether the meaning of one text expression can be inferred from another \cite{DaganGl04}. 
Given two pieces of text $(P, H)$, we say the premise $P$ \emph{entails} the hypothesis $H$ if the \emph{entirety} of $H$'s meaning can be most likely inferred true after a human reads $P$. If some units of meaning in $H$ are contradicted by, or cannot be determined 
from $P$, we describe the relation between the two as \emph{contradiction} or \emph{neutral} \cite{MarneffeRaMa08} respectively.
This fundamentally challenging natural language understanding task provides a general interface for semantic inference and comparison across different sources of textual information.

\begin{table}[t]
\small
    \centering
    \begin{tabular}{llllc}
    \toprule
    \multicolumn{5}{l}{\textbf{Premise Document}}  \\
    \midrule
    \multicolumn{5}{p{0.95\linewidth}}{Andrew Warhola, known as Andy Warhol, is an American artist born August 6, 1928 in Pittsburgh, Pennsylvania and died February 22, 1987 in New York. He is one of the main representatives of pop art. Warhol is known the world over for his work as a painter, music producer, author, avant-garde films... (7 more sentences omitted)} \\
    \midrule
    \multicolumn{5}{p{0.95\linewidth}}{\textbf{Hypothesis Sentence}} \\ \multicolumn{5}{p{0.95\linewidth}}{(\textit{from another document of the same topic})} \\
    \midrule
    \multicolumn{5}{p{0.95\linewidth}}{... The Andy Warhol Museum in his hometown, Pittsburgh, Pennsylvania, contains an extensive permanent collection of art. ...} \\
    \midrule
    \multicolumn{2}{p{0.59\linewidth}}{\textbf{Propositions}} & \multicolumn{3}{r}{\textbf{Entailment Label}} \\ 
    \cmidrule(r){1-4} \cmidrule{5-5}
    \multicolumn{4}{p{0.65\linewidth}}{\textcolor{blue}{The Andy Warhol Museum} \textcolor{lightgray}{in his hometown, Pittsburgh, Pennsylvania,} \textcolor{blue}{contains an extensive permanent collection of art}.} & \textit{Neutral} \\
    \cmidrule(r){1-4} \cmidrule{5-5}
     \multicolumn{4}{p{0.65\linewidth}}{\textcolor{lightgray}{The} \textcolor{blue}{Andy Warhol} \textcolor{lightgray}{Museum in} \textcolor{blue}{his hometown, Pittsburgh, Pennsylvania}\textcolor{lightgray}{, contains an extensive permanent collection of art.}} & \textit{\textcolor{green!60!black}{Entailment}} \\
    \cmidrule(r){1-4} \cmidrule{5-5}
     \multicolumn{4}{p{0.65\linewidth}}{\textcolor{blue}{The Andy Warhol Museum in his hometown, Pittsburgh, Pennsylvania}\textcolor{lightgray}{, contains an extensive permanent collection of art.}} & \textit{Neutral} \\
     \bottomrule
    \end{tabular}
    \caption{An example instance from the \datasetname dataset with propositions (marked as token subsets highlighted in \textcolor{blue}{blue}) and their entailment labels. }
    \label{tab:lead-example}
    \vspace{-5pt}
\end{table}

In reality, most naturally occurring text expressions are composed of a variable number of \emph{propositions}, i.e. distinct units of meaning conveyed by the piece of text. Consider the sentence shown in Table \ref{tab:lead-example}: ``\textit{The Andy Warhol Museum in his hometown, Pittsburgh, Pennsylvania, contains an extensive permanent collection of art.}'' Despite the sentence being relatively compact, it still contains (at least) three propositions, as listed in Table~\ref{tab:lead-example}. 
While the entire hypothesis would be classified as \emph{neutral} or \emph{not-entailed} to the premise,  one of its propositions ``\emph{Andy Warhol's hometown is in Pittsburgh, Pennsylvania}'' is in fact entailed by the premise, while the premise provides no support for the other two propositions. 
This phenomenon, namely \emph{partial entailment} \cite{ levy-etal-2013-recognizing}, is a blind spot for existing sentence- or paragraph-level NLI formulations. 
When a hypothesis is \textit{compositional}, NLI labels coarsely defined on the sentence/paragraph-level cannot express the difference between partial entailment from the non-entailment cases. 



This work argues for the need to study and model textual entailment relations on the level of \emph{propositions}. 
As NLI tasks and applications typically involve different genre of text with variable length and number of propositions \cite{yin-etal-2021-docnli}, decomposing textual entailment relation to the propositional level provides a more fine-grained yet accurate description of textual entailment relation between two arbitrary text expressions. Modeling \emph{propositional textual entailment} provides a more unified inference format across NLI tasks, and would potentially improve the generalization capabilities of NLI models, e.g.\ with respect to the variability in input lengths \cite{schuster2022stretching}.\looseness=-1

We propose \datasetname, a multi-domain corpus with over 45K human-annotated propositions.
\footnote{The dataset is available at \url{https://github.com/google-research-datasets/propsegment}} 
We define the tasks of proposition-level segmentation and entailment. Given a hypothesis sentence and a premise document, a system is expected to segment the hypothesis into the set of propositions, and recognize whether each proposition can be inferred from the premise. 

Interestingly, we observe that existing notions of proposition adopted by Open Information Extraction (OpenIE) or Semantic Role Labeling (SRL) \cite{baker1998berkeley, kingsbury2002treebank, meyers2004nombank} 
often fail to account for the complete set of propositions in a sentence, partly due to the fact that predicates and arguments in different propositions do not necessarily follow the same granularity (\S\ref{sec:desiderata}). 
We therefore adopt a more flexible and unified way of representing a proposition as a \textit{subset of tokens} from the input sentence, without explicitly annotating the semantic role or predicate-argument structure within the proposition, as illustrated in Table~\ref{tab:lead-example}.
We discuss the motivation and design desiderata in \S\ref{sec:desiderata}.

We construct \datasetname by sampling clusters of topically-aligned documents, i.e.\ documents focusing on the same entity or event, from \textsc{Wikipedia} \cite{schuster2022stretching} and the news domains \cite{gu2020generating}.
We train and instruct expert annotators to identify all propositions exhaustively in a document, and label the textual entailment relation of each proposition with respect to another document in the cluster, viewed as the premise.

We discuss the modeling challenges, and establish strong baselines for the segmentation and entailment tasks.   
We demonstrate the utility of our dataset and models through downstream use case studies on summary hallucination detection \cite{maynez-etal-2020-faithfulness}, and DocNLI \cite{yin-etal-2021-docnli}, through which we show that recognizing and decomposing entailment relations at the proposition-level could provide fine-grained characterization and explanation for NLI-like tasks, especially with long and compositional hypotheses.

In summary, the main contributions in our paper include: (1) Motivating the need to recognize textual entailment relation on proposition level; (2) Introducing the first large-scale dataset for studying proposition-level segmentation and entailment recognition; and (3) Leveraging \datasetname to train Seq2Seq models as strong baselines for the tasks, and demonstrating their utility in document-level NLI and hallucination detection tasks.

%% file: sections/motivation.tex
\section{Motivations \& Design Challenges}
\label{sec:desiderata}
Our study concerns the challenges of applying NLI/RTE task formulations and systems in \textit{real-world} downstream applications and settings. 
As textual entailment describes the relation between the meanings of two text expressions, 
one natural type of downstream use cases for NLI systems is to identify alignments and discrepancies between the semantic content presented in different documents/sources \cite{kryscinski-etal-2020-evaluating, schuster-etal-2021-get, chen-etal-2022-generating}. 

\begin{figure}[t]
    \centering
\includegraphics[width=0.85\linewidth]{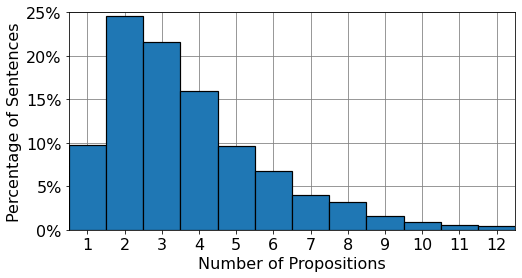}
    \vspace{-5pt}
    \caption{Distribution of proposition counts in sentences with at least one informational propositions from Wikipedia and news in \datasetname. }
    \label{fig:dist-prop}
    \vspace{-10pt}
\end{figure}

Our study is motivated by the task of comparing the content of topically-related documents, e.g.\ news documents covering the same event \cite{gu2020generating}, or Wikipedia pages from different languages for similar entities \cite{schuster2022stretching}. As existing NLI datasets typically define the textual entailment relations at the sentence or paragraph level \cite{bowman-etal-2015-large, williams-etal-2018-broad}, NLI systems trained on such resources can only recognize whether or not the entirety of a hypothesis sentence/paragraph is entailed by a premise. 
However, we estimate that, in these two domains, around $90\%$ of the sentences that convey any informational propositions contain more than one proposition (Figure~\ref{fig:dist-prop}).  In the presence of multiple propositions, \textit{partial entailment} \cite{levy-etal-2013-recognizing} describes the phenomenon where only a subset of propositions in the hypothesis is entailed by the premise.\looseness=-1


\begin{figure}[t]
    \centering
\includegraphics[width=0.85\linewidth]{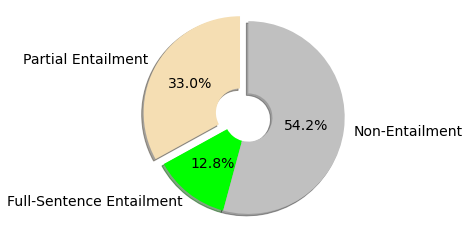}
    \vspace{-5pt}
    \caption{The percentage of sentences with \emph{partial entailment} relation to another topically-related document from Wikipedia or news in \datasetname. Typically, NLI/RTE datasets do not distinguish partial entailment from the non-entailment categories. }
    \label{fig:percentage}
    \vspace{-10pt}
\end{figure}

\paragraph{Partial entailment is $3\times$ more common than full-sentence entailment.} 
In our corpus, we observe that, given two topically related documents from news or Wikipedia, $46\%$ of sentences in one document have at least some information supported by the other document (Figure~\ref{fig:percentage}). But $74\%$ of such sentences are \textit{partially entailed}, with only some propositions supported by the other document. 
In this sense, a sentence-level NLI model can only detect a quarter of sentences that have meaningful entailment relations. In applications that seek a full understanding of cross-document semantic links, there is thus $4\times$ headroom, a significant blind spot for sentence-level NLI models.

As we observe that most natural sentences are compositional, i.e. contain more than one proposition,
we argue for the need to decompose and recognize textual entailment relation at the more granular level of propositions. In other words, instead of assessing the entire hypothesis as one unit in the context of a premise, we propose to evaluate the truth value of each proposition individually, and aggregate for the truth value of the hypothesis. 


\paragraph{Current predicate-argument based methods often fail to extract all propositions in a sentence.} 

The linguistic notion of a proposition refers to a single, contextualized unit of meaning conveyed in a sentence.
In the NLP community, propositions are usually represented by the predicate-argument structure of a sentence. For example, resources like FrameNet~\cite{baker1998berkeley}, PropBank \cite{palmer2005proposition}, NomBank \cite{meyers2004nombank}, among others, represent a proposition by a predicate (verbal, nominal, etc.), with arguments filling its thematic proto-roles. 
Such resources facilitate the development of SRL systems \cite{palmer2010semantic} for proposition extraction, with a closed, predefined set of proto-roles. 
To increase the coverage of propositions extracted, OpenIE formulations \cite{etzioni2008open, del2013clausie, cui-etal-2018-neural} were proposed to forgo the limits on fixed semantic roles and account for both explicit and implicit predicates.
However, we observe that OpenIE systems often fail to account for the complete set of propositions in a sentence. In many cases, e.g. the \textit{Andy Warhol's hometown} example in Table~\ref{tab:lead-example}, arguments of a proposition might not follow the same granularity as the ones in the sentence, e.g. \textit{Andy Warhol} vs \textit{Andy Warhol Museum}. 
Also, as OpenIE triples are still defined on direct predicate-argument relations, they often fail to produce a \textit{decontextualized} \cite{choi-etal-2021-decontextualization} view of a proposition. 
For example, an OpenIE system would recognize the possessive relation ``he has a hometown'', but fail to resolve the references of \textit{he} $\rightarrow$ \textit{Andy Warhol}, and \textit{hometown} $\rightarrow$ \textit{Pittsburgh}. 

Furthermore, \citet{gashteovski-etal-2020-aligning} and \citet {fatahi-bayat-etal-2022-compactie} observe that \textit{neural} OpenIE systems tend to extract long arguments that could potentially be decomposed into more compact propositions.
For textual entailment, we argue for the need to extract the complete set of propositions in their most \emph{compact} form, due to the fact that their truth value could vary individually.

To illustrate the difference between OpenIE and our approach, we offer a list of example propositions from our proposed \datasetname dataset, and compared them to extractions from rule-based and neural OpenIE systems, in Appendix~\ref{appendix:openie-examples}.



%% file: sections/dataset.tex
\section{\datasetname Dataset}
\label{sec:dataset}
We propose \datasetname, a large-scale dataset featuring clusters of topically similar news and Wikipedia documents, with human annotated propositions and entailment labels.  
\subsection{Task Definitions}
\label{ssec:task_def}
\begin{figure}[t]
    \centering
\includegraphics[width=0.9\linewidth]{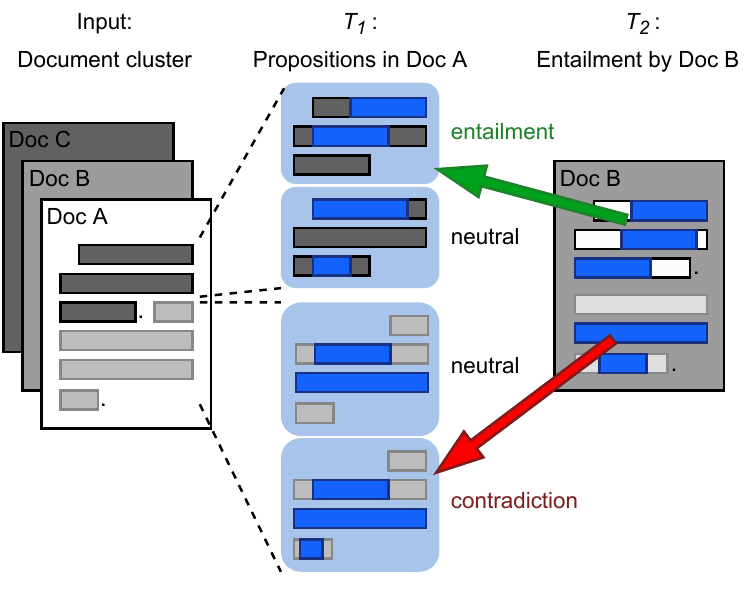}
    \vspace{-5pt}
    \caption{Given a cluster of related documents, $T_1$ asks for each sentence of each document to be segmented into propositions, represented as subsets of a sentence's tokens. $T_2$ asks to classify the entailment relation \{\emph{entails, neutral, contradicts}\} of each proposition in document A w.r.t. another document B from the same cluster; Our annotations also feature a single proposition in B that best supports each \emph{entails} or \emph{contradicts} label.}
    \label{fig:task-diagram}
    \vspace{-10pt}
\end{figure}

We formulate the task of recognizing propositional textual entailment into two sub-tasks~(Fig.~\ref{fig:task-diagram}). Given a hypothesis sentence and a premise document, a system is expected to (1) identify all the propositions within the hypothesis sentence, and (2) classify the textual entailment relation of each proposition with respect to the premise document. 

\begin{table*}[h]
\small
    \centering
    \begin{tabular}{c|ccc|ccc|ccc}
    \toprule
    \multirow{ 2}{*}{Item} & \multicolumn{3}{c|}{\textsc{Wikipedia}} & \multicolumn{3}{c|}{\textsc{News}} & \multicolumn{3}{c}{\textsc{Full Dataset}}\\
    
    & Train & Dev & Test & Train & Dev & Test & Train & Dev & Test \\
    \midrule
    News Clusters & $210$ & $15$ & $24$ & $210$ & $15$ & $25$ & $420$ & $30$ & $49$\\
    Documents & $630$ & $45$ & $72$ & $630$ & $45$ & $75$ & $1260$ & $90$ & $147$ \\
    Sentences & $4990$ & $376$ & $532$ & $4923$ & $348$ & $596$ & $9913$ & $724$ & $1128$\\
    Propositions & $21191$ & $1597$ & $2380$ & $17015$ & $1344$ & $2023$ & $38206$ & $2941$ & $4403$\\ 
    Prop.$\rightarrow$Doc. Label \# & $14083$ & $1057$ & $4729$ & $11369$ & $948$ & $ 4008$ & $25452$ & $2005$ & $8737$ \\
    \textsc{Entail} Label \% & $34.70$ & $33.24$ & $34.85$ & $20.27$ & $19.98$ & $20.13$ & $28.26$ & $26.99$ & $28.19$\\ 
     \bottomrule
    \end{tabular}
    \caption{Notable Statistics from the \datasetname dataset.}
    \label{tab:dataset-stat}
    \vspace{-8pt}
\end{table*}

    

\paragraph{$T_1$: Propositional Segmentation}
\label{ssec:task_def_prop_ext}

Given a sentence $S$ with tokens $[t_0, t_1, ..., t_l]$ from a document $D$, a system is expected to identify the set of propositions $\mathcal{P} \subseteq 2^S$, where each proposition $p \in \mathcal{P}$ is represented by a unique subset of tokens in sentence S. 
In other words, each proposition can be represented in sequence labeling format,
per the example from Table~\ref{tab:lead-example}. Each proposition is expected (1) to correspond to a distinct fact that a reader learns directly from reading the given sentence, (2) include all tokens within the sentence that are relevant to learning this fact, and (3) to not be equivalent to a conjunction of other propositions.
We opt for this format as it does not require explicit annotation of the predicate-argument structure. This allows for more expressive power for propositions with implied or implicit predicates \cite{stern-dagan-2014-recognizing}. 
Also, representing each proposition as a separate sequence could effectively account for cases with shared predicate or arguments spans, and make evaluation more readily accessible.

Since the propositions, as we demonstrated earlier, do not necessarily have a unique and identifiable predicate word in the sentence, the typical inference strategy, e.g. in SRL or OpenIE, which first extracts the set of predicates, and then identifies the arguments with respect to each predicate would not work in this case.
For this reason, given an input sentence, we expect a model on the task to directly output \emph{all} propositions. 
In such \textit{one-to-set} prediction setting, the output propositions of the model are evaluated as an unordered set. 
\paragraph{$T_2$: Propositional Entailment}
Given a hypothesis proposition $p$ from document $D_{hyp}$ and a whole premise document $D_{prem}$, a system is expected to classify whether the premise entails the proposition, i.e.\ if the information conveyed by the proposition would
be inferred true from the
premise. 



\subsection{Dataset Construction}
\label{ssec:dataset-construction}

We sample 250 document clusters from both the Wiki Clusters \cite{schuster2022stretching} and \newshead \cite{gu2020generating} datasets.
Each cluster contains the first 10 sentences of three documents, either news articles on the same event, or Wikipedia pages in different languages (machine-translated into English) of the same entity.
For each sentence, we train and instruct three human raters to annotate the set of propositions, each of which represented by a unique subset of tokens from the sentence. Conceptually, we instruct raters to include all the words that (1) pertain to the content of a proposition, and (2) are explicitly present in the sentence. For example, if there does not exist a predicate word for a proposition in the sentence, then only include the corresponding arguments. Referents present within the sentence are included in addition to pronominal and nominal references.
We provide a more detailed description of our rater guidelines and how propositions are defined with respect to various linguistic phenomena in Appendix~\ref{appendix:rater-guidelines}.

Given the three sets of propositions from the three raters for a sentence, we reconcile and select one of the three raters' responses with the highest number of propositions that the other raters also annotate. 
Since the exact selection of tokens used to mark a proposition may vary across different raters, we allow for fuzziness when measuring the match between two propositions. 
Following \citet{fitzgerald-etal-2018-large} and \citet{roit-etal-2020-controlled}, we use Jaccard similarity, i.e. intersection over union of the two sets of selected tokens, to measure the similarity between two propositions. We say two propositions match if their Jaccard similarity  is greater or equal to a threshold $\theta = 0.8$, and align two raters' responses using unweighted bipartite matching between propositions satisfying the Jaccard threshold.

Next, for all propositions in a document, we sample one other document from the document cluster as premise, and ask three raters to label the textual entailment relation between each proposition and the premise, i.e. one of \{\textit{Entailment, Neutral, Contradiction}\}. 
We take the majority vote from the three as the gold entailment label.   
Interestingly, we observe that only $0.2\%$ of all annotated labels from the rater are ``\textit{contradictions}''. We speculate that the low presence of contraditions can in part be attributed to the difficulty in establishing reference determinacy \cite{bowman-etal-2015-large} between the premise and hypothesis. We discuss more details in Appendix~\ref{appendix:contradictions}.
For this reason, we choose to only consider two-way label (\{\textit{Entailment, Non-Entailment}\}) for the entailment task evaluation. 

We create the train/dev/test splits based on clusters, so that documents in each cluster exclusively belong to only one of the splits. Overall, the dataset features 1497 documents with ${\sim}45$K propositions with entailment labels; More statistics in Table~\ref{tab:dataset-stat}. 

\subsection{Inter-Rater Agreement}
For the propositional segmentation task ($T_1$), as the inter-rater agreement involves set-to-set comparison between the propositions annotated by a pair of raters, we report two different metrics. 

First, between each pair of raters, we use the same Jaccard similarity with $\theta=0.8$ and find the matched set of propositions between the raters with bipartite matching for each example. 
We measure the coverage of the matched set by either rater with $F_1$ score. We observe 0.57 $F_1$ among all raters. As comparison, we use the same metric for model evaluation and human performance estimation, as we will discuss in \S~\ref{ssec:evaluation}.
In addition, we measure the token-level agreement on the matched set of propositions among raters with Fleiss' kappa \cite{fleiss1971measuring}, i.e. whether raters agree on whether each token should be included in a proposition or not. We observed $\kappa=0.63$, which indicates moderate to substantial agreement among raters.

For the entailment task, ($T_2$), we observe Fleiss' kappa = $0.84$ across three-way \{\textit{Entailment, Neutral, Contradiction}\} labels.

%% file: sections/baseline.tex
\begin{table*}[t]
\small
    \centering
    \begin{tabular}{cccccccc}
    \toprule
    \multirow{ 2}{*}{Task/Setting} & \multirow{ 2}{*}{Model} & \multicolumn{3}{c}{Jaccard $\theta=0.8$} & \multicolumn{3}{c}{Exact Match} \\
     &  & Precision & Recall & F1 & Precision & Recall & F1 \\
    \cmidrule(r){1-1} \cmidrule(lr){2-2} \cmidrule(lr){3-5} \cmidrule(l){6-8}
    
    \multirow{ 7}{*}{\parbox{0.15\linewidth}{\centering\textsc{T}$_1$: Propositional Segmentation}} & \textsc{Bert}-Base & $33.77$ & $33.53$ & $33.65$ & $14.33$ & $14.60$ & $14.47$\\ 
    & \textsc{Bert}-Large & $34.97$ & $33.42$ & $34.17$ & $14.61$ & $14.16$ & $14.38$ \\ 
    & \textsc{T5}-Base & $54.96$ & $51.93$ & $53.41$ & $32.87$ & $31.54$ & $32.19$\\
    & \textsc{T5}-Base \emph{w/ Entail.} & $53.54$ & $51.50$ & $52.50$ & $31.61$ & $30.67$ & $31.13$ \\ 
     &   \textsc{T5}-Large & $\textbf{55.95}$ & $\textbf{55.05}$ & $\textbf{55.50}$ & $\textbf{32.40}$ & $\textbf{32.16}$ & $\textbf{32.28}$ \\
    & \textsc{T5}-Large \emph{w/ Entail.} & $\textbf{56.27}$ & $\textbf{55.50}$ & $\textbf{55.89}$ & $\textbf{31.94}$ & $\textbf{32.11}$ & $\textbf{32.02}$ \\
    \cmidrule(lr){2-2} \cmidrule(lr){3-5} \cmidrule(l){6-8}
    & Human Performance & $69.63$ & $64.69$ & $67.07$ & $44.86$	& $42.93$ &	$43.87$ \\
    \midrule
    \multirow{7}{*}{\parbox{0.15\linewidth}{\centering \textsc{T}$_2$: Propositional Entailment}} &  & \multicolumn{3}{c}{Performance (2-way Class.)} & \multicolumn{3}{c}{Per-Label $F_1$ (3-way Class.)}\\
     &  & Accuracy & \multicolumn{2}{c}{Balanced Accuracy} & Entail. & Neutral & Contra. \\
    \cmidrule(lr){2-2} \cmidrule(lr){3-5} \cmidrule(l){6-8}
    & \textit{Always Entails.} & $27.89$ & \multicolumn{2}{c}{$50.00$} & $43.62$ & $0.00$ & $\textcolor{lightgray}{0.00}$ \\
    & \textit{Always Neutral} & $72.10$ & \multicolumn{2}{c}{$50.00$} & $0.00$ & $83.54$ & $\textcolor{lightgray}{0.00}$ \\
    & \textsc{T5}-Base & $85.17$ & \multicolumn{2}{c}{$81.44$} &  $73.32$ & $89.68$ & $\textcolor{lightgray}{11.21}$\\
    & \textsc{T5}-Large & $\textbf{91.38}$ & \multicolumn{2}{c}{\textbf{$\textbf{89.75}$}} & $\textbf{84.78}$ & $\textbf{93.98}$ & $\textcolor{lightgray}{\textbf{20.34}}$ \\
    \cmidrule(lr){2-2} \cmidrule(lr){3-5} \cmidrule(l){6-8}
    & Human Performance & $90.20$ & \multicolumn{2}{c}{\textbf{$88.31$}} & - & - & -\\ 
     \bottomrule
    \end{tabular}
    \caption{Performance of the baseline models on the full (\textsc{Wiki} + \textsc{News}) test set. Due to the low presence of contradictions 
    ($32 / 8643 = 0.4\%$ of test), $F_1$ for contradiction does not reflect statistically significant improvement.}
    \label{tab:baseline-results}
    \vspace{-8pt}
\end{table*}

\section{Baseline Methods}
\subsection{Propositional Segmentation Baselines}
The key challenge with the proposition extraction task lies within the one-to-set structured prediction setting.
Our one-to-set prediction format is similar to QA-driven semantic parsing such as QA-SRL \cite{he-etal-2015-question, klein2022qasem}, as both involve generating a variable number of units of semantic content under no particular order between them. 
As in propositions, there might not necessarily be a unique and identifiable predicate word associated with each proposition, extracting predicates first (e.g. as a sequence tagging task), and later individually produce one proposition for each predicate would not be a sufficient solution in this case.  

For this particular one-to-set problem setup, 
We introduce two classes of baseline models. 
\paragraph{Seq2Seq: T5} \cite{raffel2020exploring} When formatting a output set as a sequence, Seq2Seq models have been found to be a strong method for tasks with set outputs, as they employ chain-rules to efficiently model the joint probability of outputs \cite{Vinyals2016OrderMS}. 
The obvious caveat for representing set outputs as sequences is that we need an ordering for the outputs. 
Having a consistent ordering helps seq2seq model learn to maintain the output set structure \cite{Vinyals2016OrderMS}, and the best ordering scheme is often both model- and task-specific \cite{klein2022qasem}. In our experiments, we observe that sorting the propositions by the appearance order of the tokens in the sentence, i.e. positions of the foremost tokens of each proposition in the sentence, yields the best performance. 

We start from the pretrained T5 1.1 checkpoints from the T5x library \cite{roberts2022scaling}. Given a sentence input, we finetune the T5 model to output the propositions in a single sequence. 
For each input sentence, we sort the output propositions using the aforementioned ordering scheme, and join them by a special token \texttt{[TARGET]}. The spans of tokens included in each proposition is surrounded by special tokens \texttt{[M]} and \texttt{[/M]}.
For instance, \textit{``\markstart Alice\markend~and Bob~\markstart went to the Zoo\markend.~\texttt{[TARGET]} Alice and \markstart Bob went to the Zoo.\markend~''}. 
In addition, we evaluate the setting where the model is also given the premise document $D_{prem}$, and learns to output the entailment label along with each proposition (T5 \textit{w/ Entail.} in Table~\ref{tab:baseline-results}). 

\paragraph{Encoder+Tagger: BERT} \cite{devlin-etal-2019-bert} For comparison, we provide a simpler baseline that does not model joint probability of the output propositions. On top of the last layer an encoder model, i.e. BERT, we add $k$ linear layers that each correspond to one output proposition. 
Given an input sentence, the $i^{th}$ linear layer produces a binary ($0/1$) label per token, indicating whether the token is in the $i^{th}$ proposition or not. 
$k$ is set to be a sufficiently large number, e.g. $k=20$ in our experiments. 
We use the label of the \texttt{[CLS]} token of the $i^{th}$ linear layer to indicate whether the $i^{th}$ proposition should exist in the output. For such, we follow the same ordering of the output propositions as in the seq2seq (T5) baseline setup.   
\subsection{Propositional Entailment Baselines}
\label{ssec:entailment-baseline}
We formulate the task as a sequence labeling problem,
and finetune T5 model as our baseline. The inputs consist of the hypothesis proposition $p$ with its document context $D_{hyp}$, plus the premise document $D_{prem}$. The output is one of the three-way labels \{\textit{Entailment, Neutral, Contradiction}\}. 
Due to low presence of contradictions, we merge the \textit{neutral} and \textit{contradiction} outputs from the model as \textit{non-entailments} during evaluation. To ensure that the model has access to the essential context information, our task input also include the document $D_{hyp}$ of the hypothesis proposition $p$, so that model has a decontextualized view of $p$ when inferring its textual entailment relation with $D_{prem}$.

%% file: sections/experiments.tex
\section{Experiments and Results}
\label{sec:experiments}

\subsection{Evaluation Metrics}
\label{ssec:evaluation}

\paragraph{Propositional Segmentation}
We measure the precision and recall between the set of predicted and gold propositions for a given sentence. 
As the set of gold propositions do not follow any particular ordering, we first produce a bipartite matching between them using the Hungarian algorithm \cite{kuhn1955hungarian}. 
We again use the Jaccard similarity over $\theta = 0.8$ as a fuzzy match between two propositions (\S~\ref{ssec:dataset-construction}).
We also use exact match, an even more restrictive measure where two propositions match if and only if they have the exact same tokens. We report the macro-averaged precision and recall over sentences in the test set. 

\paragraph{Propositional Entailment}
We report the baseline performance under two-way classification results in accuracy. We also report the balanced accuracy, i.e. average of true positive and true negative rate, due to label imbalance (Table~\ref{tab:dataset-stat}). To understand the per-label performance, we also report the $F_1$ score w.r.t. each of the three-way label. 

\begin{table}[t]
\small
    \centering
    \begin{tabular}{c|c|cc}
    \toprule
    \multirow{ 5}{*}{\rotatebox[origin=c]{90}{Train Domain}} &  \multicolumn{3}{c}{Test Domain} \\
    & \multicolumn{3}{c}{($P/R/F_1$ \emph{w/ Jaccard} $\theta = 0.8$ )}\\
    \multicolumn{2}{c}{} & \textsc{Wiki} & \textsc{News} \\
    \cmidrule(lr){2-4}
     & \textsc{Wiki} & $53.95/53.16/53.56$ & $44.93/44.95/44.94$ \\ 
    & \textsc{News} & $45.21/43.65/44.42$ &  $49.58/47.81/48.68$\\ 
     \bottomrule
    \end{tabular}
    \caption{Cross-domain (i.e. train on \textsc{News} $\rightarrow$ test on \textsc{Wiki}, and train on \textsc{Wiki} $\rightarrow$ test on \textsc{News}) generalization results of \textsc{T5}-large on the segmentation (\textsc{T}$_1$) task. }
    \label{tab:cross-domain-results}
    \vspace{-10pt}
\end{table}

\subsection{Baseline Results}
Table~\ref{tab:baseline-results} shows the evaluation results for the segmentation (\textsc{T}$_1$) and entailment task (\textsc{T}$_2$) respectively.  

For the segmentation task (\textsc{T}$_1$), the seq2seq T5 model setup yields superior performance compared to the simpler encoder+tagger BERT setup.
As the encoder+tagger setup predicts each proposition individually, and does not attend on other propositions during inference, we observe that the model predicts repeated/redundant propositions in $>20\%$ of the input sentences. In the seq2seq T5 setup, the repetition rate is $<1\%$. 
For both setups, we remove the redundant outputs as a post processing step.
We also evaluate the multi-task setup (i.e. T5 \textit{w/ Entail.} in Table~\ref{tab:baseline-results}) where the model jointly learns the entailment label with each proposition, and observe no significant improvements. 
For the entailment task (\textsc{T}$_2$), we see that T5-Large yields the best overall performance. 
We observe that the performance with respect to the \emph{entailment} label is lower compared to the \emph{neutral} label. 

For both tasks, we estimate the averaged human expert performance by comparing annotations from three of the authors to ground truth on 50 randomly sampled examples from the dataset. 
We observe that for the segmentation task $T_1$, we observe that the human performance increases after reconciling and selecting the ground truth response ($0.57 \rightarrow 0.67~F_1$). We see that there remains a sizable gap between the best model, T5-Large, and human performance. On the entailment task $T_2$, T5-Large exceeds human performance, which is not uncommon among language inference tasks of similar classification formats \cite{wang2019glue}.

\subsection{Cross-Domain Generalization}
On the propositional segmentation (\textsc{T}$_1$) task, we evaluate the how the best baseline model generalizes across the Wikipedia (\texttt{Wiki}) and \texttt{News} domains. Table~\ref{tab:cross-domain-results} shows the results of T5-Large models finetuned on data from each domain, and evaluated on the test split of both domains.

When applying a model trained on \texttt{Wiki}, we see a larger drop in performance when tested on \texttt{News}, as the \texttt{News} domain features more syntactic and stylistic variations compared to the \texttt{Wiki} domain.

%% file: sections/discussion.tex
\section{Analysis and Discussion}
We exemplify the utilities of our propositional segmentation and entailment framework, which we refer to as PropNLI, through the lens of two downstream use cases, e.g. summary hallucination detection (\S~\ref{ssec:hallucination}), and document-level NLI w/ variable-length hypotheses (\S~\ref{ssec:docnli}).

\begin{table}[t]
\small
    \centering
    \begin{tabular}{p{0.95\linewidth}}
    \toprule
\textbf{Document:} The incident happened near Dr Gray's Hospital shortly after 10:00.
The man was taken to the hospital with what police said were serious but not life-threatening injuries.
The A96 was closed in the area for several hours, but it has since reopened. \\
     \midrule
     \textbf{Summary w/ human labeled \underline{hallucinated spans}:} \\ 
     A man has been taken to hospital following a \underline{one-vehicle crash} on the A96 \underline{in Aberdeenshire}.\\
     
     \midrule
     \textbf{Predicted propositions (\textcolor{blue}{blue}) and entailment labels} \\ 
     \textit{\#1:} \textcolor{blue}{A man has been taken to hospital} \textcolor{lightgray}{following a one-vehicle crash on the A96 in Aberdeenshire.} \greencheck \\
     \textit{\#2:} \textcolor{blue}{A man has been taken to hospital following a one-vehicle crash} \textcolor{lightgray}{on the A96 in Aberdeenshire}. \redmark \\
     \textit{\#3:} \textcolor{lightgray}{A man has been taken to hospital following a} \textcolor{blue}{one-vehicle crash on the A96} \textcolor{lightgray}{in Aberdeenshire}. \redmark \\
     \textit{\#4:} \textcolor{lightgray}{A man has been taken to hospital following a} \textcolor{blue}{one-vehicle crash} \textcolor{lightgray}{on the A96} \textcolor{blue}{in Aberdeenshire}. \redmark \\
     \midrule
     \textbf{Predicted hallucinated spans} (union of \redmark - union of \greencheck)\\
     A man has been taken to hospital \underline{following a} \underline{one-vehicle crash on the A96 in Aberdeenshire}. \\
     \bottomrule
    \end{tabular}
    \caption{An example model generated summary on the XSum dataset, with human-annotated hallucination spans from \citet{maynez-etal-2020-faithfulness}. We show that we can infer the hallucinated spans from the set of four propositions and their entailment labels (\textit{entail}=\greencheck, \textit{not-entail}=\redmark), predicted by our T5-Large models. More examples can be found in Appendix~\ref{appendix:xsum-examples}}
    \label{tab:xsum-example}
    \vspace{-5pt}
\end{table}

\begin{table}[t]
\small
    \centering
    \begin{tabular}{c|c|cccccc}
    \toprule
    \multirow{3}{*}{Method} & Hallu. & \multicolumn{6}{c}{Span Detection} \\
    & Class. & \multicolumn{3}{c}{\textit{Faith. Tokens}} & \multicolumn{3}{c}{\textit{Hallu. Tokens}} \\
    & \textit{B. Acc.} & $P$ & $R$ & $F_1$ & $P$ & $R$ & $F_1$\\
     \midrule
     PropNLI & $\textbf{.62}$ & .78 & .50 & .61 & .64 & .71 & .67 \\
     MNLI & $.59$ & .96 & .17 & .30 & .56 & .88 & .68 \\
     \bottomrule
    \end{tabular}
    \caption{Zero-shot performance of PropNLI vs. T5-Large MNLI model on hallcination identification and span detection tasks from \citet{maynez-etal-2020-faithfulness}.}
    \label{tab:xsum-results}
    \vspace{-5pt}
\end{table}

\subsection{Application: Hallucination Detection}
\label{ssec:hallucination} 
We first look at the task of summary hallucination detection, i.e. given a summary of a source document, identify whether the summary's content is \emph{faithful} to the document. Naturally the task can be represented as a NLI problem, and NLI systems have been shown effective on the task \cite{kryscinski-etal-2020-evaluating, chen-etal-2021-improving}. As summaries can be long and compositional, recognizing partial entailment, and identifying which part(s) of a summary is hallucinated becomes important \cite{goyal-durrett-2020-evaluating, laban2022summac}. 

To show that PropNLI can be used for hallucination detection, we experiment on the model generated summaries on the XSum dataset \cite{narayan-etal-2018-dont}, where \citet{maynez-etal-2020-faithfulness} provide human annotations of the sets of hallucinated spans (if they exist) in the summaries. 
Table~\ref{tab:xsum-example} illustrates our idea.
If a proposition in a summary is \emph{entailed} by the document, then all spans covered by the proposition are faithful. 
Otherwise, \textit{some} spans would likely contain \textit{hallucinated} information.

Following such intuitions, we first evaluate the performance of our method in zero-shot settings as a hallucination classifier , i.e. binary classification for whether a summary is hallucinated or not.  
For baseline comparison, we use a T5-large model finetuned on MNLI \cite{williams-etal-2018-broad} to classify a full summary as entailed ($\rightarrow$ \textit{faithful}) or not ($\rightarrow$ \textit{hallucinated}). As $\tilde~89\%$ of the summaries annotated by \citet{maynez-etal-2020-faithfulness} are hallucinated, we again adopt balanced accuracy (\S~\ref{ssec:evaluation}) as the metric. 
On 2500 examples, our method achieved $61.68\%$ balanced accuracy, while MNLI achieved $58.79\%$.

Next, we study whether the entailment labels of propositions can be composed to detect hallucinated spans in a summary. 
As in Table~\ref{tab:xsum-example}, we take the union of the spans in \emph{non-entailed} propositions, and exclude the spans that has appeared in \textit{entailed} propositions. 
The intuition is that the hallucinated information likely only exists in the non-entailed propositions
, but not the entailed ones.

We evaluate hallucinated span detection as a token classification task. For each summary, we evaluate the precision and recall of the \textit{faithful} and \textit{hallucinated} set of predicted tokens respectively against the human-labeled ground truth set. We report the macro-averaged precision, recall and $F_1$ score over all 2,500 summaries. 
We compare our method to a T5-Large model finetuned on MNLI, where we label all tokens as \emph{faithful} if the summary is predicted to be \emph{entailed}, and all tokens as \emph{hallucinated} otherwise. 
We report the performance with respect to each of the two labels in Table~\ref{tab:xsum-results}.
As the MNLI model don't distinguish partial entailment from non-entailment cases, it predicts more tokens to be hallucinated, and thus having low precision and high recall on the hallucinated tokens, and vice versa.   
On the other hand, we observe our model can be used to detect the nuance between faithful and hallucinated tokens with good and more balanced performance for both cases. 
Table~\ref{tab:xsum-example} shows one example summary and PropNLI's predictions, 
and we include more examples in Appendix~\ref{appendix:xsum-examples}.

\subsection{Proposition-Level $\rightarrow$ Sentence/Paragraph-Level Entailment}
\label{ssec:docnli}
We would like to see whether proposition-level entailment labels can potentially be \emph{composed} to explain sentence/paragraph-level NLI predictions. 

Given a hypothesis sentence/paragraph and a premise, our PropNLI framework takes three steps. First we segment the hypothesis into propositions. For each proposition, we infer its entailment relation with the premise. 
In cases where multiple propositions exist in the hypothesis, the proposition-level entailment labels can be aggregated to obtain the entailment label for the \emph{entire} hypothesis, similar to ideas presented in \citet{stacey2022logical}. 
As a starting point, we assume logical conjunction as the aggregation function, and hypothesize that this will offer a more fine-grained and explainable way of conducting NLI inference. 



\begin{figure}[t]
    \centering
\includegraphics[width=0.8\linewidth]{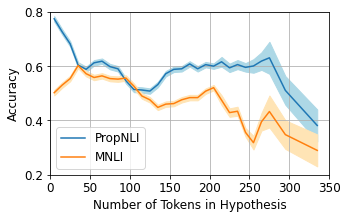}
    \vspace{-5pt}
        \caption{Zero-shot performance of T5-large MNLI model compared to our PropNLI T5-large models (i.e. segmentation `$\rightarrow$ entailment $\rightarrow$ aggregation) with respect to varying \emph{hypothesis length} in DocNLI dev set. The shaded region shows 95\% confidence interval. }
    \label{fig:docnli}
    \vspace{-10pt}
\end{figure}

To demonstrate the utility of the idea, we conduct a case study on DocNLI \cite{yin-etal-2021-docnli}, which features premise and hypothesis of different length, and so varying number and compositions of propositions. 
We take the baseline T5-Large segmentation and entailment models respectively, and use logical conjunction to aggregate the proposition-level entailment prediction. 
We compare PropNLI in a zero-shot setting against the T5-Large MNLI model. The MNLI model takes the entire hypothesis and premise and input without any segmentation or decomposition.

The results are shown in Figure~\ref{fig:docnli}. We take the development set of DocNLI and split examples into buckets according to number of tokens in the hypothesis. We examine the zero-shot performance of the PropNLI setup versus the finetuned MNLI model.  
We observe that with shorter hypotheses ($<100$ tokens), the two setups demonstrated similar performance, as the hypothesis length is similar to the distribution of MNLI training set (avg. $21.73$ tokens $\pm 30.70$). 
As the length of the hypothesis increases, the performance of MNLI model starts to drop, while PropNLI's performance remains relatively stable. 
Such observations suggest the potential of using the PropNLI framework to describe the textual entailment relations between a pair of premise and hypothesis in a more precise and fine-grained manner. 
In the realistic case where input hypotheses are compositional, the \datasetname present an opportunity for developing more generalizable NLI models and solutions.

%% file: sections/conclusion.tex
\section{Conclusion}
In this paper, we presented \datasetname, the first large-scale dataset for studying proposition-level segmentation and entailment. 
We demonstrate that segmenting a text expression into propositions, i.e. atomic units of meanings, and assessing their truth values would provide a finer-grained characterization of the textual entailment relation between two pieces of text. 
Beyond NLI/RTE tasks, we hypothesize that proposition-level segmentation might be helpful in similar ways for other text classification tasks as well. 
We hope that \datasetname will serve as a starting point, and pave a path for research forward along the line.

%% file: sections/limitations.tex
\section*{Limitations}
Since the \datasetname dataset feature entailment labels for \textit{all} propositions in a document, the label distribution are naturally imbalanced, which would potentially pose challenge for modeling. We observe low presence of contradiction examples in our dataset construction process, which could be a limiting factor for the utility of the dataset.
Unlike previous NLI datasets~\cite{bowman-etal-2015-large, williams-etal-2018-broad}, we speculate that reference determinacy, i.e. whether the hypothesis and premise refer to the same scenario at the same time, cannot be certainly guaranteed and safely assumed in our case, which in part leads to low presence of contradictions during annotation.
We offer a detailed discussion on the implications of reference determinacy and contradictions in Appendix~\ref{appendix:contradictions}. We leave the exploration on \emph{natural} contradictions for future work. 

As the annotation complexity and cost scales quadratically w.r.t. the number of propositions in a document, we truncate the documents in \datasetname to the first ten sentences of the original document. 

\section*{Ethical Considerations}
In the proposition-level entailment task ($T_2$), the inference of the entailment relation between a premise document and a hypothesis proposition uses the \emph{assumption} that the premise document is true. The assumption is common to NLI datasets \cite{dagan2005pascal, bowman-etal-2015-large, williams-etal-2018-broad}, and is necessary for the task's structure. With the documents in \datasetname, we make the assumption only for the experimental purpose of $T_2$, and make no claim about the actual veracity of the premise documents.

%% file: sections/acknowledgements.tex
\section*{Acknowledgements}
We thank Michael Collins, Corinna Cortes, Paul Haahr, Ilya Kornakov, Ivan Kuznetsov, Annie Louis, Don Metzler, Jeremiah Milbauer, Pavel Nalivayko, Fernando Pereira, Sandeep Tata, Yi Tay, Andrew Tomkins, and Victor Zaytsev for insightful discussions, suggestions, and support. We are grateful to the annotators for their work in creating \datasetname.

%% file: appendix/model_param.tex
\section{Model Implementation}

\paragraph{T5} We use T5 1.1 checkpoints from the T5x library \cite{roberts2022scaling}, with Flaxformer\footnote{\url{https://github.com/google/flaxformer}} implementation. 
For all sizes of T5 model and all tasks, we finetune the model for three epoch, with $1e-3$ learning rate, $0.1$ dropout rate, batch size of $128$. We train the models on 16 TPU v3 slices. 

\paragraph{BERT} We use the BERT English uncased models from Tensorflow \cite{abadi2016tensorflow}, in large (24 layers, 16 attention heads, 1024 max sequence length) and base (12 layers, 12 attention heads, 768 max sequence length) sizes. For both sizes, we finetune the model for five epoch, with $1e-5$ learning rate, $0.1$ dropout rate, batch size of $16$.  We train the models on 8 TPU v3 slices. 

%% file: appendix/rater_guidelines.tex
\section{Annotation Guidelines}
\label{appendix:rater-guidelines}

\subsection{Segmentation annotation guidelines}

There is no unequivocally unique definition for precisely how to segment an English sentence in the context of a document into propositions defined as token subsets, due to a variety of complex language phenomena. Our raters were instructed to follow the following overall guidelines for the segmentation task:

\begin{enumerate}
    \item Each proposition is expected to correspond to a distinct fact that a reader learns directly from reading the given sentence.
    \begin{enumerate}
        \item The raters are instructed to focus on the text's most literal \textit{denotation}, rather than drawing further inferences from the text based on world knowledge, external knowledge, or common sense.
        \item The raters are instructed to consider \textit{factivity}, marking only propositions that, in their judgement, the author intends the reader to take as factual from reading the sentence.
        \item With regard to quotes, raters are asked to estimate the author's intent, including the proposition quoted when the reader is expected to take it as factual, and/or the proposition of the quote itself having been uttered if the reader is expected to learn that a speaker uttered that quote.
        \item The raters are instructed to omit text that are clearly non-factual, such as rhetorical flourishes or first-person account of an article author's emotional response to the topic. This rule is specific to the news and Wikipedia domains, since in other domains of prose, first-person emotions may well be part of the intended informational payload.
    \end{enumerate}
    \item Each proposition should include all tokens within the sentence that are relevant to learning this fact.
    \begin{enumerate}
        \item Specifically, the raters are asked to include any tokens in the same sentence that are antecedents of pronouns or other endophora in the proposition, or relevant bridging references.
        \item Raters are asked to ignore punctuation, spacing, and word inflections when selecting tokens, though a number of other minutiae, such as whether to include articles, are left unspecified in the rater instructions.
    \end{enumerate}
    \item Choose the simplest possible propositions, so that no proposition is equivalent to a conjunction of the other propositions, and so that the union of all of the sentence's proposition gives us all the information a reader learns from the sentence.
\end{enumerate}

The raters are also asked to omit propositions from any text that doesn't constitute well-formed sentences, typically arising from parsing errors or from colloquialisms.

Note that the resulting subsets of tokens do not, generally, constitute well-formed English sentences when concatenated directly, but can, in our ad hoc trials, easily be reconstituted into stand-alone sentences by a human reader.

\subsection{Entailment annotation guidelines}

For the propositional entailment task, our instructions are somewhat similar to the RTE task \cite{DaganGl04}, but specialized to the proposition level.

The raters are asked to read the premise document and decide whether a specific hypothesis proposition is entailed by it, contradicted, or neither. In the first two cases, the raters are asked to mark a proposition in the premise document that most closely supports the hypothesis proposition, using the same definition of proposition as above. The interface nudges the raters to select one of the propositions marked by the segmentation rater, but allows the entailment rater to create a new proposition as well. Note that the choice of a specific supporting proposition is sometimes not well defined.

To judge entailment, the raters are asked ``from reading just the premise document, do we learn that the hypothesis proposition is true, learn that it's false, or neither?'' More specifically, the raters are asked:

\begin{enumerate}
    \item To consider the full document of the hypothesis as the context of the hypothesis proposition, and the full premise document.
    \item To allow straightforward entailment based on ``common sense or widely-held world knowledge'', but otherwise
    avoid entailment labels whenever ``significant analysis'' (any complex reasoning, specialized knowledge, or subjective judgement) is required to align the two texts.
    \item To assume that the two documents were written in the same coarse spatiotemporal context --- same geographical area, and the same week.
\end{enumerate}

Raters have the option of marking that they don't understand the premise and/or the hypothesis and skipping the question.

%% file: appendix/contradictions.tex
\section{Reference Determinacy and Contradictions}
\label{appendix:contradictions}
The \datasetname dataset is constructed in document-to-document comparison settings. Even though the document clusters are sampled so that documents in a cluster target the same event or event, the documents typically have different focus. Besides the factual information, which are mostly consistent across documents, the focus or specific perspective of each document varies largely, which is in part why we observe very few contradictions. 

Apart from such, We speculate that the low presence of contradictions can also be in part attributed to the difficulty in establishing reference determinacy, i.e. whether the entities and events described in a hypothesis can be assumed to refer to the same ones or happening at the same point in the premise. To illustrate the importance of this, consider the following example from SNLI \cite{bowman-etal-2015-large}.

\begin{quote}
    \textit{Premise}: A black race car starts up in front of a crowd of people. \\
    \textit{Hypothesis}: A man is driving down a lonely road. 
\end{quote}

In SNLI, reference determinacy is assumed to be true. In other words, the human raters assume that the scenario described in the premise and hypothesis happens in the same context at the same time point.  Therefore, the example pair is labeled as contradiction, as ``lonely road'' contradicts ``a crowd of people'' if we assume both happen on the same road. Without such assumption, the example would likely be labeled as \textit{neutral}, since there is no extra context that would indicate the two events happen in the same context. 

In reality, reference determinacy is often difficult to establish with certainty. Unlike existing NLI/RTE datasets \cite{dagan2005pascal, bowman-etal-2015-large, williams-etal-2018-broad}, in the creation process of \datasetname, we do not assume reference determinacy between the hypothesis proposition and premise document, but rather relay the judgement to human raters by reading context information presented in the documents. 
We observe that it is often hard to tell if a specific proposition within a document can establish reference determinacy with the other document, unless the proposition describes a property that is stationary with respect to time. 
For this reason, most contradictions, among the few that exist in our dataset, are factual statements. Here is an example from the development split. 

\begin{quote}
    \textit{Premise}: ... The team was founded in 1946 as a founding member of the All-America Football Conference (AAFC) and joined the NFL in 1949 when the leagues merged.. \\
    \textit{Hypothesis}: \textcolor{lightgray}{The 49ers have been members of the NFL since} \textcolor{blue}{the AAFC and National Football League} \textcolor{lightgray}{(NFL)} \textcolor{blue}{merged in 1950}...
\end{quote}

We view the lack of contradictions as a potential limitation for the dataset for practical purposes. We argue for the need to circumscribe the exact definition of contradiction (from the practical perspective) when reference determinacy cannot be simply assumed. We leave this part for future work.

%% file: appendix/openie_vs_us.tex
\section{Example Propositions From OpenIE vs. \datasetname}
\label{appendix:openie-examples}
To illustrate the difference between how we define propositions in \datasetname, versus OpenIE formulations, we include a few examples sentences with propositions in \datasetname in Table~\ref{tab:openie-examples-appendix} and \ref{tab:openie-examples-cont-appendix},  and compare propositions extracted with ClausIE, a rule-based OpenIE model \cite{del2013clausie}, and a neural Bi-LSTM model from \citet{stanovsky-etal-2018-supervised}. 
\begin{table*}[h]
\small
    \centering
    \begin{tabular}{p{0.95\linewidth}}
    \toprule
\textbf{Sentence:} The 82nd NFL Draft took place from April 27-29, 2017 in Philadelphia. \\
\textbf{\datasetname} \\
\#1: \textcolor{blue}{The 82nd NFL Draft took place from April 27-29, 2017} in Philadelphia. \\
\#2: \textcolor{blue}{The 82nd NFL Draft took place} from April 27-29, 2017 \textcolor{blue}{in Philadelphia}. \\
\textbf{ClausIE}  \\
\#1: (The 82nd NFL Draft, took place, from April 27-29, 2017 in Philadelphia) \\
\#2: (The 82nd NFL Draft, took place, from April 27-29, 2017) \\
\textbf{Neural Bi-LSTM OIE} \textit{(Splitting each modifier, i.e. ARGM)} \\
\#1: (The 82nd NFL Draft, took, place, from April 27-29, 2017) \\
\#2: (The 82nd NFL Draft, took, place, in Philadelphia) \\

\midrule
\textbf{Sentence:} She has also appeared in films such as Little Women (1994), The Hours (2002), Self Defense (1997), Les Miserables (1998) and Orson Welles y yo (2009). \\
\textbf{\datasetname} \\
\#1: \textcolor{blue}{She has also appeared in films such as Little Women} (1994), The Hours (2002), Self Defense (1997), Les Miserables (1998) and Orson Welles y yo (2009). \\
\#2: \textcolor{blue}{She has also appeared in films such as} Little Women (1994), \textcolor{blue}{The Hours} (2002), Self Defense (1997), Les Miserables (1998) and Orson Welles y yo (2009). \\
\#3: \textcolor{blue}{She has also appeared in films such as} Little Women (1994), The Hours (2002), \textcolor{blue}{Self Defense} (1997), Les Miserables (1998) and Orson Welles y yo (2009). \\
\#4: \textcolor{blue}{She has also appeared in films such as} Little Women (1994), The Hours (2002), Self Defense (1997), \textcolor{blue}{Les Miserables} (1998) and Orson Welles y yo (2009). \\
\#5: \textcolor{blue}{She has also appeared in films such as} Little Women (1994), The Hours (2002), Self Defense (1997), Les Miserables (1998) and \textcolor{blue}{Orson Welles y yo} (2009). \\
\#6: She has also appeared in films such as \textcolor{blue}{Little Women (1994)}, The Hours (2002), Self Defense (1997), Les Miserables (1998) and Orson Welles y yo (2009). \\
\#7: She has also appeared in films such as Little Women (1994), \textcolor{blue}{The Hours (2002)}, Self Defense (1997), Les Miserables (1998) and Orson Welles y yo (2009). \\
\#8: She has also appeared in films such as Little Women (1994), The Hours (2002), \textcolor{blue}{Self Defense (1997)}, Les Miserables (1998) and Orson Welles y yo (2009). \\
\#9: She has also appeared in films such as Little Women (1994), The Hours (2002), Self Defense (1997), \textcolor{blue}{Les Miserables (1998)} and Orson Welles y yo (2009). \\
\#10: She has also appeared in films such as Little Women (1994), The Hours (2002), Self Defense (1997), Les Miserables (1998) and \textcolor{blue}{Orson Welles y yo} (2009). \\
\textbf{ClausIE}  \\
\#1: (She, has appeared, in films such as Little Women also) \\
\#2: (She, has appeared, in films such as The Hours also) \\
\#3: (She, has appeared, in films such as Self Defense also) \\
\#4: (She, has appeared, in films such as Les Miserables also) \\
\#5: (She, has appeared, in films such as Orson Welles y yo also) \\
\#6: (She, has appeared, in films such as Little Women) \\
\#7: (She, has appeared, in films such as The Hours) \\
\#8: (She, has appeared, in films such as Self Defense) \\
\#9: (She, has appeared, in films such as Les Miserables) \\
\#10: (She, has appeared, in films such as Orson Welles y yo) \\
\#11: (Little Women, is, 1994) \\
\#12: (The Hours, is, 1994) \\
\#13: (Self Defense, is, 1994) \\
\#14: (Les Miserables, is, 1994) \\
\#15: (Orson Welles y yo, is, 1994) \\
\#16: (The Hours, is, 2002) \\
\#17: (Self Defense, is, 1997) \\
\#18: (Les Miserables, is, 1998) \\
\#19: (Orson Welles y yo, is, 2009) \\
\textbf{Neural Bi-LSTM OIE} \\
\#1: (She, appeared, in films such as Little Women (1994), The Hours (2002), Self Defense (1997), Les Miserables (1998) and Orson Welles y yo (2009)) \\
     \bottomrule
    \end{tabular}
    \caption{Comparison of propositions in \datasetname with extractions with ClausIE \cite{del2013clausie}, and the neural Bi-LSTM OIE model from \citet{stanovsky-etal-2018-supervised}.}
    \label{tab:openie-examples-appendix}
\end{table*}

\begin{table*}[h]
\small
    \centering
    \begin{tabular}{p{0.95\linewidth}}
    \toprule
    \textbf{Sentence:} The Andy Warhol Museum in his hometown, Pittsburgh, Pennsylvania, contains an extensive permanent collection of art. \\
    \textbf{\datasetname} \\
    \#1: \textcolor{blue}{The Andy Warhol Museum} in his hometown, Pittsburgh, Pennsylvania, \textcolor{blue}{contains an extensive permanent collection of art}. \\
    \#2: The \textcolor{blue}{Andy Warhol} Museum in \textcolor{blue}{his hometown, Pittsburgh, Pennsylvania}, contains an extensive permanent collection of art. \\
    \#3: \textcolor{blue}{The Andy Warhol Museum in his hometown, Pittsburgh, Pennsylvania}, contains an extensive permanent collection of art. \\
    \textbf{ClausIE}  \\
    \#1: (his, has, hometown) \\
    \#2: (his hometown, is, Pittsburgh Pennsylvania) \\
    \#3: (The Andy Warhol Museum in his hometown, contains, an extensive permanent collection of art) \\ 
    \textbf{Neural Bi-LSTM OIE} \\
    \#1: (The Andy Warhol Museum in his hometown Pittsburgh Pennsylvania, contains, an extensive permanent collection of art) \\
    \midrule
\textbf{Sentence:} The Cleveland Cavaliers got the first choice in the lottery, which was used on 20-year-old forward Anthony Bennett, a freshman from the University of Nevada. \\
\textbf{\datasetname} \\
\#1: \textcolor{blue}{The Cleveland Cavaliers got the first choice in the lottery}, which was used on 20-year-old forward Anthony Bennett, a freshman from the University of Nevada. \\
\#2: \textcolor{blue}{The Cleveland Cavaliers got} the first choice in the lottery, which was used on 20-year-old forward \textcolor{blue}{Anthony Bennett}, a freshman from the University of Nevada. \\
\#3: The Cleveland Cavaliers got the first choice in the lottery, which was used on \textcolor{blue}{20-year-old} forward \textcolor{blue}{Anthony Bennett}, a freshman from the University of Nevada. \\
\#4: The Cleveland Cavaliers got the first choice in the lottery, which was used on 20-year-old \textcolor{blue}{forward Anthony Bennett}, a freshman from the University of Nevada. \\
\#5: The Cleveland Cavaliers got the first choice in the lottery, which was used on 20-year-old forward \textcolor{blue}{Anthony Bennett, a freshman} from the University of Nevada. \\
\#6: The Cleveland Cavaliers got the first choice in the lottery, which was used on 20-year-old forward \textcolor{blue}{Anthony Bennett}, a freshman \textcolor{blue}{from the University of Nevada}. \\
\textbf{ClausIE}  \\
\#1: (The Cleveland Cavaliers, got, the first choice in the lottery) \\
\#2: (the lottery, was used, on 20-year-old forward Anthony Bennett) \\
\#3: (Anthony Bennett, is, a freshman from the University of Nevada) \\
\textbf{Neural Bi-LSTM OIE} \\
\#1: (The Cleveland Cavaliers, got, the first choice in the lottery, which was used on 20-year-old forward Anthony Bennett, a freshman from the University of Nevada.)\\
\#2: (the lottery, was used, on 20-year-old forward Anthony Bennett, a freshman from the University of Nevada.)\\

\midrule
\bottomrule
    \end{tabular}
    \caption{(Cont.) Comparison of propositions in \datasetname with extractions with ClausIE \cite{del2013clausie}, and the neural Bi-LSTM OIE model from \citet{stanovsky-etal-2018-supervised}.}
    \label{tab:openie-examples-cont-appendix}
\end{table*}

%% file: appendix/xsum_examples.tex
\section{XSum Hallucination Detection - Examples}
\label{appendix:xsum-examples}
Table~\ref{tab:xsum-example-appendix} and~\ref{tab:xsum-example-appendix-cont} show two example documents, with propositions and the inferred hallucinated spans in model-generated and gold summaries by our PropNLI model. We compare the predictions to the annotations of hallucinated span provided by \citet{maynez-etal-2020-faithfulness}.

\begin{table*}[t]
\small
    \centering
    \begin{tabular}{p{0.95\linewidth}}
    \toprule
\textbf{Document:} The incident happened near Dr Gray's Hospital shortly after 10:00.
The man was taken to the hospital with what police said were serious but not life-threatening injuries.
The A96 was closed in the area for several hours, but it has since reopened. \\
     \midrule
     \textbf{Summary from \texttt{BertS2S}} \\ 
     A man has been taken to hospital following a \underline{one-vehicle crash} on the A96 \underline{in Aberdeenshire}.\\
     
     \textbf{Predicted propositions (\textcolor{blue}{blue}) and entailment labels} \\ 
     \textit{\#1:} \textcolor{blue}{A man has been taken to hospital} following a one-vehicle crash on the A96 in Aberdeenshire. \greencheck \\
     \textit{\#2:} \textcolor{blue}{A man has been taken to hospital following a one-vehicle crash} on the A96 in Aberdeenshire. \redmark \\
     \textit{\#3:} A man has been taken to hospital following a \textcolor{blue}{one-vehicle crash on the A96} in Aberdeenshire. \redmark \\
     \textit{\#4:} A man has been taken to hospital following a \textcolor{blue}{one-vehicle crash} on the A96 \textcolor{blue}{in Aberdeenshire}. \redmark \\
     \textbf{Predicted hallucinated spans} (union of \redmark - union of \greencheck)\\
     A man has been taken to hospital \underline{following a one-vehicle crash on the A96 in Aberdeenshire}. \\
     \midrule
     \textbf{Summary from \texttt{TConvS2S}} \\ 
     a man has been taken to hospital \underline{after being hit by a car in Moray}. \\
     \textbf{Predicted propositions (\textcolor{blue}{blue}) and entailment labels} \\ 
     \textit{\#1:} \textcolor{blue}{a man has been taken to hospital} after being hit by a car in Moray.  \greencheck\\
     \textit{\#2:} \textcolor{blue}{a man has been taken to hospital after being hit by a car in Moray}. \redmark \\
     \textbf{Predicted hallucinated spans} (union of \redmark - union of \greencheck)\\
     a man has been taken to hospital \underline{after being hit by a car in Moray}. \\ 
     \midrule
     
     \textbf{Gold Summary from the XSum dataset} \\
     \underline{A cyclist has suffered serious head injuries after a collision with a car in Elgin}. \\
     \textbf{Predicted propositions (\textcolor{blue}{blue}) and entailment labels} \\ 
     \textit{\#1:} \textcolor{blue}{A cyclist has suffered serious head injuries} after a collision with a car in Elgin. \redmark \\
     \textit{\#2:} \textcolor{blue}{A cyclist has suffered serious head injuries after a collision with a car} in Elgin. \redmark \\
     \textit{\#3:} \textcolor{blue}{A cyclist has suffered serious head injuries} after a collision with a car \textcolor{blue}{in Elgin}. \redmark \\
     \textbf{Predicted hallucinated spans} (union of \redmark - union of \greencheck) \\
     \underline{A cyclist has suffered serious head injuries after a collision with a car in Elgin}. \\
     \midrule
     
     \textbf{Summary from \texttt{PTGen}} \\ 
     A man has been taken to hospital after being \underline{hit by a car} in the A96 area of \underline{Glasgow}. \\
     \textbf{Predicted propositions (\textcolor{blue}{blue}) and entailment labels} \\ 
     \textit{\#1:} \textcolor{blue}{A man has been taken to hospital} after being hit by a car in the A96 area of Glasgow. \greencheck \\
     \textit{\#2:} \textcolor{blue}{A man has been taken to hospital after being hit by a car} in the A96 area of Glasgow. \redmark \\
     \textit{\#3:} \textcolor{blue}{A man} has been taken to hospital after being \textcolor{blue}{hit by a car in the A96 area of Glasgow}. \redmark \\
     \textbf{Predicted hallucinated spans} (union of \redmark - union of \greencheck) \\
     A man has been taken to hospital  \underline{after being hit by a car in the A96 area of Glasgow} \\
     \midrule
     
     \textbf{Summary from \texttt{TranS2S}} \\ 
     A man has been taken to hospital after a \underline{two-vehicle crash} on the A96 \underline{in County Antrim}. \\
     \textbf{Predicted propositions (\textcolor{blue}{blue}) and entailment labels} \\ 
     \textit{\#1:} \textcolor{blue}{A man has been taken to hospital} after a two-vehicle crash on the A96 in County Antrim. \greencheck \\
     \textit{\#2:} \textcolor{blue}{A man has been taken to hospital after a two-vehicle crash} on the A96 in County Antrim. \redmark \\
     \textit{\#3:} \textcolor{blue}{A man has been taken to hospital after a two-vehicle crash on the A96} in County Antrim. \redmark \\
     \textit{\#4:} \textcolor{blue}{A man has been taken to hospital} after a two-vehicle crash on the A96 \textcolor{blue}{in County Antrim}. \redmark \\
     \textbf{Predicted hallucinated spans} (union of \redmark - union of \greencheck) \\
     A man has been taken to hospital \underline{after a two-vehicle crash on the A96 in County Antrim}. \\
     \bottomrule
    \end{tabular}
    \caption{More example of model generated summaries on the XSum dataset, with human-annotated hallucination spans from \citet{maynez-etal-2020-faithfulness}. For each document, \citet{maynez-etal-2020-faithfulness} provide summaries and hallucination annotations from 5 different summarization systems. We randomly sample documents and show our model's predictions for all 5 summaries here. }
    \label{tab:xsum-example-appendix}
    \vspace{-5pt}
\end{table*}

\begin{table*}[t]
\small
    \centering
    \begin{tabular}{p{0.95\linewidth}}
    \toprule
\textbf{Document:} Dervite, 28, made 14 appearances last season to help Wanderers finish second in League One and secure promotion.
The French centre-back joined Bolton from Charlton in 2014 and has made 83 appearances in all competitions.
"Dorian was a bit of a forgotten man last year but came in and made an excellent contribution towards the end of the campaign," manager Phil Parkinson told the club website.
Dervite follows David Wheater, Gary Madine and Jem Karacan in signing new contracts with Bolton, following their promotion to the Championship.\\
     \midrule
     \textbf{Summary from \texttt{BertS2S}} \\ 
     Bolton defender Dorian Dervite has signed a new \underline{two-year} contract with the championship club.\\
     
     \textbf{Predicted propositions (\textcolor{blue}{blue}) and entailment labels} \\ 
     \textit{\#1:} \textcolor{blue}{Bolton defender Dorian Dervite} has signed a new \underline{two-year} contract with the championship club. \greencheck \\
     \textit{\#2:} Bolton defender \textcolor{blue}{Dorian Dervite has signed a new two-year contract with the championship club}. \redmark \\
     \textbf{Predicted hallucinated spans} (union of \redmark - union of \greencheck)\\
     Bolton defender Dorian Dervite \underline{has signed a new two-year contract with the championship club}. \\
     \midrule
     \textbf{Summary from \texttt{TConvS2S}} \\ 
     Bolton Wanderers have signed \underline{defender} Dorian Dervite \underline{from bolton wanderers} for an \underline{undisclosed fee}. \\
     \textbf{Predicted propositions (\textcolor{blue}{blue}) and entailment labels} \\ 
     \textit{\#1:} \textcolor{blue}{Bolton Wanderers} have \textcolor{blue}{signed} defender \textcolor{blue}{Dorian Dervite} from bolton wanderers for an undisclosed fee. \redmark\\
     \textit{\#2:} \textcolor{blue}{Bolton Wanderers} have \textcolor{blue}{signed} defender \textcolor{blue}{Dorian Dervite from bolton wanderers} for an undisclosed fee. \redmark \\
     \textit{\#3:} \textcolor{blue}{Bolton Wanderers} have \textcolor{blue}{signed} defender \textcolor{blue}{Dorian Dervite} from bolton wanderers \textcolor{blue}{for an undisclosed fee}. \redmark \\
     \textit{\#4:} Bolton Wanderers have signed \textcolor{blue}{defender Dorian Dervite} from bolton wanderers for an undisclosed fee. \greencheck\\
     \textbf{Predicted hallucinated spans} (union of \redmark - union of \greencheck)\\
     \underline{Bolton Wanderers} have \underline{signed} defender Dorian Dervite \underline{from bolton wanderers for an undisclosed fee}. \\  
     \midrule
     
     \textbf{Gold Summary from the XSum dataset} \\
     \underline{Defender Dorian Dervite has signed a new one-year contract with Bolton}. \\
     \textbf{Predicted propositions (\textcolor{blue}{blue}) and entailment labels} \\ 
     \textit{\#1:} \textcolor{blue}{Defender Dorian Dervite} has signed a new one-year contract with Bolton \greencheck \\
     \textit{\#2:} Defender \textcolor{blue}{Dorian Dervite has signed a new one-year contract with Bolton}. \redmark \\
     \textbf{Predicted hallucinated spans} (union of \redmark - union of \greencheck) \\
     Defender Dorian Dervite \underline{has signed a new one-year contract with Bolton}. \\
     \midrule
     
     \textbf{Summary from \texttt{PTGen}} \\ 
     Bolton Wanderers defender Dorian Dervite has signed a \underline{new three-and-a-half-year contract with the league one club} \underline{until the end of the 2018-19 season}. \\
     \textbf{Predicted propositions (\textcolor{blue}{blue}) and entailment labels} \\ 
     \textit{\#1:} \textcolor{blue}{Bolton Wanderers defender Dorian Dervite} has signed a new three-and-a-half-year contract with the league one club until the end of the 2018-19 season. \greencheck \\
     \textit{\#2:} Bolton Wanderers defender \textcolor{blue}{Dorian Dervite has signed a new three-and-a-half-year contract with the league one club} until the end of the 2018-19 season. \redmark \\
     \textit{\#3:} Bolton Wanderers defender \textcolor{blue}{Dorian Dervite has signed a new three-and-a-half-year contract} with the league one club \textcolor{blue}{until the end of the 2018-19 season}. \redmark \\
     \textbf{Predicted hallucinated spans} (union of \redmark - union of \greencheck) \\
     Bolton Wanderers defender Dorian Dervite \underline{has signed a new three-and-a-half-year contract with the league one club} \underline{until the end of the 2018-19 season}. \\
     \midrule
     
     \textbf{Summary from \texttt{TranS2S}} \\ 
     \underline{Bolton Wanderers midfielder Gary Wheat has signed a new one-year contract with the championship side}. \\
     \textbf{Predicted propositions (\textcolor{blue}{blue}) and entailment labels} \\ 
     \textit{\#1:} \textcolor{blue}{Bolton Wanderers midfielder Gary Wheat} has signed a new one-year contract with the championship side. \redmark \\
     \textit{\#2:} Bolton Wanderers midfielder \textcolor{blue}{Gary Wheat has signed a new one-year contract with the championship side}.  \redmark \\
     \textbf{Predicted hallucinated spans} (union of \redmark - union of \greencheck) \\
     \underline{Bolton Wanderers midfielder Gary Wheat has signed a new one-year contract with the championship side}. \\
     \bottomrule
    \end{tabular}
    \caption{(Cont.) More example of model generated summaries on the XSum dataset, with human-annotated hallucination spans from \citet{maynez-etal-2020-faithfulness}.}
    \label{tab:xsum-example-appendix-cont}
    \vspace{-5pt}
\end{table*}

%% file: main.bbl
\begin{thebibliography}{43}
\expandafter\ifx\csname natexlab\endcsname\relax\def\natexlab#1{#1}\fi

\bibitem[{Abadi et~al.(2016)Abadi, Barham, Chen, Chen, Davis, Dean, Devin,
  Ghemawat, Irving, Isard et~al.}]{abadi2016tensorflow}
Martin Abadi, Paul Barham, Jianmin Chen, Zhifeng Chen, Andy Davis, Jeffrey
  Dean, Matthieu Devin, Sanjay Ghemawat, Geoffrey Irving, Michael Isard, et~al.
  2016.
\newblock {TensorFlow}: a system for {Large-Scale} machine learning.
\newblock In \emph{12th USENIX symposium on operating systems design and
  implementation (OSDI 16)}, pages 265--283.

\bibitem[{Baker et~al.(1998)Baker, Fillmore, and Lowe}]{baker1998berkeley}
Collin~F Baker, Charles~J Fillmore, and John~B Lowe. 1998.
\newblock The {Berkeley} {FrameNet} project.
\newblock In \emph{COLING 1998 Volume 1: The 17th International Conference on
  Computational Linguistics}.

\bibitem[{Bowman et~al.(2015)Bowman, Angeli, Potts, and
  Manning}]{bowman-etal-2015-large}
Samuel~R. Bowman, Gabor Angeli, Christopher Potts, and Christopher~D. Manning.
  2015.
\newblock \href {https://doi.org/10.18653/v1/D15-1075} {A large annotated
  corpus for learning natural language inference}.
\newblock In \emph{Proceedings of the 2015 Conference on Empirical Methods in
  Natural Language Processing}, pages 632--642, Lisbon, Portugal. Association
  for Computational Linguistics.

\bibitem[{Chen et~al.(2022)Chen, Sriram, Choi, and
  Durrett}]{chen-etal-2022-generating}
Jifan Chen, Aniruddh Sriram, Eunsol Choi, and Greg Durrett. 2022.
\newblock \href {https://aclanthology.org/2022.emnlp-main.229} {Generating
  literal and implied subquestions to fact-check complex claims}.
\newblock In \emph{Proceedings of the 2022 Conference on Empirical Methods in
  Natural Language Processing}, pages 3495--3516, Abu Dhabi, United Arab
  Emirates. Association for Computational Linguistics.

\bibitem[{Chen et~al.(2021)Chen, Zhang, Sone, and
  Roth}]{chen-etal-2021-improving}
Sihao Chen, Fan Zhang, Kazoo Sone, and Dan Roth. 2021.
\newblock \href {https://doi.org/10.18653/v1/2021.naacl-main.475} {Improving
  faithfulness in abstractive summarization with contrast candidate generation
  and selection}.
\newblock In \emph{Proceedings of the 2021 Conference of the North American
  Chapter of the Association for Computational Linguistics: Human Language
  Technologies}, pages 5935--5941, Online. Association for Computational
  Linguistics.

\bibitem[{Choi et~al.(2021)Choi, Palomaki, Lamm, Kwiatkowski, Das, and
  Collins}]{choi-etal-2021-decontextualization}
Eunsol Choi, Jennimaria Palomaki, Matthew Lamm, Tom Kwiatkowski, Dipanjan Das,
  and Michael Collins. 2021.
\newblock \href {https://doi.org/10.1162/tacl_a_00377} {Decontextualization:
  Making sentences stand-alone}.
\newblock \emph{Transactions of the Association for Computational Linguistics},
  9:447--461.

\bibitem[{Cui et~al.(2018)Cui, Wei, and Zhou}]{cui-etal-2018-neural}
Lei Cui, Furu Wei, and Ming Zhou. 2018.
\newblock \href {https://doi.org/10.18653/v1/P18-2065} {Neural open information
  extraction}.
\newblock In \emph{Proceedings of the 56th Annual Meeting of the Association
  for Computational Linguistics (Volume 2: Short Papers)}, pages 407--413,
  Melbourne, Australia. Association for Computational Linguistics.

\bibitem[{Dagan and Glickman(2004)}]{DaganGl04}
I.~Dagan and O.~Glickman. 2004.
\newblock Probabilistic textual entailment: Generic applied modeling of
  language variability.
\newblock In \emph{Learning Methods for Text Understanding and Mining}.

\bibitem[{Dagan et~al.(2005)Dagan, Glickman, and Magnini}]{dagan2005pascal}
Ido Dagan, Oren Glickman, and Bernardo Magnini. 2005.
\newblock The {PASCAL} recognising textual entailment challenge.
\newblock In \emph{Machine learning challenges workshop}, pages 177--190.
  Springer.

\bibitem[{de~Marneffe et~al.(2008)de~Marneffe, Rafferty, and
  Manning}]{MarneffeRaMa08}
Marie-Catherine de~Marneffe, Anna~N. Rafferty, and Christopher~D. Manning.
  2008.
\newblock \href {https://aclanthology.org/P08-1118} {Finding contradictions in
  text}.
\newblock In \emph{Proceedings of ACL-08: HLT}, pages 1039--1047, Columbus,
  Ohio. Association for Computational Linguistics.

\bibitem[{Del~Corro and Gemulla(2013)}]{del2013clausie}
Luciano Del~Corro and Rainer Gemulla. 2013.
\newblock {ClausIE}: clause-based open information extraction.
\newblock In \emph{Proceedings of the 22nd international conference on World
  Wide Web}, pages 355--366.

\bibitem[{Devlin et~al.(2019)Devlin, Chang, Lee, and
  Toutanova}]{devlin-etal-2019-bert}
Jacob Devlin, Ming-Wei Chang, Kenton Lee, and Kristina Toutanova. 2019.
\newblock \href {https://doi.org/10.18653/v1/N19-1423} {{BERT}: Pre-training of
  deep bidirectional transformers for language understanding}.
\newblock In \emph{Proceedings of the 2019 Conference of the North {A}merican
  Chapter of the Association for Computational Linguistics: Human Language
  Technologies, Volume 1 (Long and Short Papers)}, pages 4171--4186,
  Minneapolis, Minnesota. Association for Computational Linguistics.

\bibitem[{Etzioni et~al.(2008)Etzioni, Banko, Soderland, and
  Weld}]{etzioni2008open}
Oren Etzioni, Michele Banko, Stephen Soderland, and Daniel~S Weld. 2008.
\newblock Open information extraction from the web.
\newblock \emph{Communications of the ACM}, 51(12):68--74.

\bibitem[{Fatahi~Bayat et~al.(2022)Fatahi~Bayat, Bhutani, and
  Jagadish}]{fatahi-bayat-etal-2022-compactie}
Farima Fatahi~Bayat, Nikita Bhutani, and H.~Jagadish. 2022.
\newblock \href {https://doi.org/10.18653/v1/2022.naacl-main.65}
  {{C}ompact{IE}: Compact facts in open information extraction}.
\newblock In \emph{Proceedings of the 2022 Conference of the North American
  Chapter of the Association for Computational Linguistics: Human Language
  Technologies}, pages 900--910, Seattle, United States. Association for
  Computational Linguistics.

\bibitem[{FitzGerald et~al.(2018)FitzGerald, Michael, He, and
  Zettlemoyer}]{fitzgerald-etal-2018-large}
Nicholas FitzGerald, Julian Michael, Luheng He, and Luke Zettlemoyer. 2018.
\newblock \href {https://doi.org/10.18653/v1/P18-1191} {Large-scale {QA}-{SRL}
  parsing}.
\newblock In \emph{Proceedings of the 56th Annual Meeting of the Association
  for Computational Linguistics (Volume 1: Long Papers)}, pages 2051--2060,
  Melbourne, Australia. Association for Computational Linguistics.

\bibitem[{Fleiss(1971)}]{fleiss1971measuring}
Joseph~L Fleiss. 1971.
\newblock Measuring nominal scale agreement among many raters.
\newblock \emph{Psychological bulletin}, 76(5):378.

\bibitem[{Gashteovski et~al.(2020)Gashteovski, Gemulla, Kotnis, Hertling, and
  Meilicke}]{gashteovski-etal-2020-aligning}
Kiril Gashteovski, Rainer Gemulla, Bhushan Kotnis, Sven Hertling, and Christian
  Meilicke. 2020.
\newblock \href {https://doi.org/10.18653/v1/2020.eval4nlp-1.14} {On aligning
  {O}pen{IE} extractions with knowledge bases: A case study}.
\newblock In \emph{Proceedings of the First Workshop on Evaluation and
  Comparison of NLP Systems}, pages 143--154, Online. Association for
  Computational Linguistics.

\bibitem[{Goyal and Durrett(2020)}]{goyal-durrett-2020-evaluating}
Tanya Goyal and Greg Durrett. 2020.
\newblock \href {https://doi.org/10.18653/v1/2020.findings-emnlp.322}
  {Evaluating factuality in generation with dependency-level entailment}.
\newblock In \emph{Findings of the Association for Computational Linguistics:
  EMNLP 2020}, pages 3592--3603, Online. Association for Computational
  Linguistics.

\bibitem[{Gu et~al.(2020)Gu, Mao, Han, Liu, Wu, Yu, Finnie, Yu, Zhai, and
  Zukoski}]{gu2020generating}
Xiaotao Gu, Yuning Mao, Jiawei Han, Jialu Liu, You Wu, Cong Yu, Daniel Finnie,
  Hongkun Yu, Jiaqi Zhai, and Nicholas Zukoski. 2020.
\newblock Generating representative headlines for news stories.
\newblock In \emph{Proceedings of The Web Conference 2020}, pages 1773--1784.

\bibitem[{He et~al.(2015)He, Lewis, and Zettlemoyer}]{he-etal-2015-question}
Luheng He, Mike Lewis, and Luke Zettlemoyer. 2015.
\newblock \href {https://doi.org/10.18653/v1/D15-1076} {Question-answer driven
  semantic role labeling: Using natural language to annotate natural language}.
\newblock In \emph{Proceedings of the 2015 Conference on Empirical Methods in
  Natural Language Processing}, pages 643--653, Lisbon, Portugal. Association
  for Computational Linguistics.

\bibitem[{Kingsbury and Palmer(2002)}]{kingsbury2002treebank}
Paul~R Kingsbury and Martha Palmer. 2002.
\newblock From {TreeBank} to {PropBank}.
\newblock In \emph{LREC}, pages 1989--1993.

\bibitem[{Klein et~al.(2022)Klein, Hirsch, Eliav, Pyatkin, Caciularu, and
  Dagan}]{klein2022qasem}
Ayal Klein, Eran Hirsch, Ron Eliav, Valentina Pyatkin, Avi Caciularu, and Ido
  Dagan. 2022.
\newblock {QASem} parsing: Text-to-text modeling of {QA}-based semantics.
\newblock \emph{arXiv preprint arXiv:2205.11413}.

\bibitem[{Kryscinski et~al.(2020)Kryscinski, McCann, Xiong, and
  Socher}]{kryscinski-etal-2020-evaluating}
Wojciech Kryscinski, Bryan McCann, Caiming Xiong, and Richard Socher. 2020.
\newblock \href {https://doi.org/10.18653/v1/2020.emnlp-main.750} {Evaluating
  the factual consistency of abstractive text summarization}.
\newblock In \emph{Proceedings of the 2020 Conference on Empirical Methods in
  Natural Language Processing (EMNLP)}, pages 9332--9346, Online. Association
  for Computational Linguistics.

\bibitem[{Kuhn(1955)}]{kuhn1955hungarian}
Harold~W Kuhn. 1955.
\newblock The {H}ungarian method for the assignment problem.
\newblock \emph{Naval research logistics quarterly}, 2(1-2):83--97.

\bibitem[{Laban et~al.(2022)Laban, Schnabel, Bennett, and
  Hearst}]{laban2022summac}
Philippe Laban, Tobias Schnabel, Paul~N Bennett, and Marti~A Hearst. 2022.
\newblock {SummaC}: Re-visiting {NLI}-based models for inconsistency detection
  in summarization.
\newblock \emph{Transactions of the Association for Computational Linguistics},
  10:163--177.

\bibitem[{Levy et~al.(2013)Levy, Zesch, Dagan, and
  Gurevych}]{levy-etal-2013-recognizing}
Omer Levy, Torsten Zesch, Ido Dagan, and Iryna Gurevych. 2013.
\newblock \href {https://aclanthology.org/P13-2080} {Recognizing partial
  textual entailment}.
\newblock In \emph{Proceedings of the 51st Annual Meeting of the Association
  for Computational Linguistics (Volume 2: Short Papers)}, pages 451--455,
  Sofia, Bulgaria. Association for Computational Linguistics.

\bibitem[{Maynez et~al.(2020)Maynez, Narayan, Bohnet, and
  McDonald}]{maynez-etal-2020-faithfulness}
Joshua Maynez, Shashi Narayan, Bernd Bohnet, and Ryan McDonald. 2020.
\newblock \href {https://doi.org/10.18653/v1/2020.acl-main.173} {On
  faithfulness and factuality in abstractive summarization}.
\newblock In \emph{Proceedings of the 58th Annual Meeting of the Association
  for Computational Linguistics}, pages 1906--1919, Online. Association for
  Computational Linguistics.

\bibitem[{Meyers et~al.(2004)Meyers, Reeves, Macleod, Szekely, Zielinska,
  Young, and Grishman}]{meyers2004nombank}
Adam Meyers, Ruth Reeves, Catherine Macleod, Rachel Szekely, Veronika
  Zielinska, Brian Young, and Ralph Grishman. 2004.
\newblock The {NomBank} project: An interim report.
\newblock In \emph{Proceedings of the workshop frontiers in corpus annotation
  at hlt-naacl 2004}, pages 24--31.

\bibitem[{Narayan et~al.(2018)Narayan, Cohen, and
  Lapata}]{narayan-etal-2018-dont}
Shashi Narayan, Shay~B. Cohen, and Mirella Lapata. 2018.
\newblock \href {https://doi.org/10.18653/v1/D18-1206} {Don{'}t give me the
  details, just the summary! topic-aware convolutional neural networks for
  extreme summarization}.
\newblock In \emph{Proceedings of the 2018 Conference on Empirical Methods in
  Natural Language Processing}, pages 1797--1807, Brussels, Belgium.
  Association for Computational Linguistics.

\bibitem[{Palmer et~al.(2005)Palmer, Gildea, and
  Kingsbury}]{palmer2005proposition}
Martha Palmer, Daniel Gildea, and Paul Kingsbury. 2005.
\newblock The proposition bank: An annotated corpus of semantic roles.
\newblock \emph{Computational linguistics}, 31(1):71--106.

\bibitem[{Palmer et~al.(2010)Palmer, Gildea, and Xue}]{palmer2010semantic}
Martha Palmer, Daniel Gildea, and Nianwen Xue. 2010.
\newblock Semantic role labeling.
\newblock \emph{Synthesis Lectures on Human Language Technologies},
  3(1):1--103.

\bibitem[{Raffel et~al.(2020)Raffel, Shazeer, Roberts, Lee, Narang, Matena,
  Zhou, Li, and Liu}]{raffel2020exploring}
Colin Raffel, Noam Shazeer, Adam Roberts, Katherine Lee, Sharan Narang, Michael
  Matena, Yanqi Zhou, Wei Li, and Peter~J Liu. 2020.
\newblock Exploring the limits of transfer learning with a unified text-to-text
  transformer.
\newblock \emph{Journal of Machine Learning Research}, 21:1--67.

\bibitem[{Roberts et~al.(2022)Roberts, Chung, Levskaya, Mishra, Bradbury,
  Andor, Narang, Lester, Gaffney, Mohiuddin et~al.}]{roberts2022scaling}
Adam Roberts, Hyung~Won Chung, Anselm Levskaya, Gaurav Mishra, James Bradbury,
  Daniel Andor, Sharan Narang, Brian Lester, Colin Gaffney, Afroz Mohiuddin,
  et~al. 2022.
\newblock Scaling up models and data with t5x and seqio.
\newblock \emph{arXiv preprint arXiv:2203.17189}.

\bibitem[{Roit et~al.(2020)Roit, Klein, Stepanov, Mamou, Michael, Stanovsky,
  Zettlemoyer, and Dagan}]{roit-etal-2020-controlled}
Paul Roit, Ayal Klein, Daniela Stepanov, Jonathan Mamou, Julian Michael,
  Gabriel Stanovsky, Luke Zettlemoyer, and Ido Dagan. 2020.
\newblock \href {https://doi.org/10.18653/v1/2020.acl-main.626} {Controlled
  crowdsourcing for high-quality {QA}-{SRL} annotation}.
\newblock In \emph{Proceedings of the 58th Annual Meeting of the Association
  for Computational Linguistics}, pages 7008--7013, Online. Association for
  Computational Linguistics.

\bibitem[{Schuster et~al.(2022)Schuster, Chen, Buthpitiya, Fabrikant, and
  Metzler}]{schuster2022stretching}
Tal Schuster, Sihao Chen, Senaka Buthpitiya, Alex Fabrikant, and Donald
  Metzler. 2022.
\newblock \href {https://aclanthology.org/2022.findings-emnlp.28} {Stretching
  sentence-pair {NLI} models to reason over long documents and clusters}.
\newblock In \emph{Findings of the Association for Computational Linguistics:
  EMNLP 2022}, pages 394--412, Abu Dhabi, United Arab Emirates. Association for
  Computational Linguistics.

\bibitem[{Schuster et~al.(2021)Schuster, Fisch, and
  Barzilay}]{schuster-etal-2021-get}
Tal Schuster, Adam Fisch, and Regina Barzilay. 2021.
\newblock \href {https://doi.org/10.18653/v1/2021.naacl-main.52} {Get your
  vitamin {C}! robust fact verification with contrastive evidence}.
\newblock In \emph{Proceedings of the 2021 Conference of the North American
  Chapter of the Association for Computational Linguistics: Human Language
  Technologies}, pages 624--643, Online. Association for Computational
  Linguistics.

\bibitem[{Stacey et~al.(2022)Stacey, Minervini, Dubossarsky, and
  Rei}]{stacey2022logical}
Joe Stacey, Pasquale Minervini, Haim Dubossarsky, and Marek Rei. 2022.
\newblock Logical reasoning with span predictions: Span-level logical atoms for
  interpretable and robust nli models.
\newblock In \emph{The Conference on Empirical Methods in Natural Language
  Processing (EMNLP)}.

\bibitem[{Stanovsky et~al.(2018)Stanovsky, Michael, Zettlemoyer, and
  Dagan}]{stanovsky-etal-2018-supervised}
Gabriel Stanovsky, Julian Michael, Luke Zettlemoyer, and Ido Dagan. 2018.
\newblock \href {https://doi.org/10.18653/v1/N18-1081} {Supervised open
  information extraction}.
\newblock In \emph{Proceedings of the 2018 Conference of the North {A}merican
  Chapter of the Association for Computational Linguistics: Human Language
  Technologies, Volume 1 (Long Papers)}, pages 885--895, New Orleans,
  Louisiana. Association for Computational Linguistics.

\bibitem[{Stern and Dagan(2014)}]{stern-dagan-2014-recognizing}
Asher Stern and Ido Dagan. 2014.
\newblock \href {https://doi.org/10.3115/v1/P14-2120} {Recognizing implied
  predicate-argument relationships in textual inference}.
\newblock In \emph{Proceedings of the 52nd Annual Meeting of the Association
  for Computational Linguistics (Volume 2: Short Papers)}, pages 739--744,
  Baltimore, Maryland. Association for Computational Linguistics.

\bibitem[{Vinyals et~al.(2016)Vinyals, Bengio, and Kudlur}]{Vinyals2016OrderMS}
Oriol Vinyals, Samy Bengio, and Manjunath Kudlur. 2016.
\newblock Order matters: Sequence to sequence for sets.
\newblock In \emph{Proceedings of the International Conference on Learning
  Representations}.

\bibitem[{Wang et~al.(2019)Wang, Singh, Michael, Hill, Levy, and
  Bowman}]{wang2019glue}
Alex Wang, Amanpreet Singh, Julian Michael, Felix Hill, Omer Levy, and Samuel~R
  Bowman. 2019.
\newblock Glue: A multi-task benchmark and analysis platform for natural
  language understanding.
\newblock In \emph{7th International Conference on Learning Representations,
  ICLR 2019}.

\bibitem[{Williams et~al.(2018)Williams, Nangia, and
  Bowman}]{williams-etal-2018-broad}
Adina Williams, Nikita Nangia, and Samuel Bowman. 2018.
\newblock \href {https://doi.org/10.18653/v1/N18-1101} {A broad-coverage
  challenge corpus for sentence understanding through inference}.
\newblock In \emph{Proceedings of the 2018 Conference of the North {A}merican
  Chapter of the Association for Computational Linguistics: Human Language
  Technologies, Volume 1 (Long Papers)}, pages 1112--1122, New Orleans,
  Louisiana. Association for Computational Linguistics.

\bibitem[{Yin et~al.(2021)Yin, Radev, and Xiong}]{yin-etal-2021-docnli}
Wenpeng Yin, Dragomir Radev, and Caiming Xiong. 2021.
\newblock \href {https://doi.org/10.18653/v1/2021.findings-acl.435}
  {{D}oc{NLI}: A large-scale dataset for document-level natural language
  inference}.
\newblock In \emph{Findings of the Association for Computational Linguistics:
  ACL-IJCNLP 2021}, pages 4913--4922, Online. Association for Computational
  Linguistics.

\end{thebibliography}
